\documentclass[lettersize,journal]{IEEEtran}
\usepackage{amsmath,amsfonts}
\usepackage{algorithmic}
\usepackage{algorithm}
\usepackage{array}
\usepackage[caption=false,font=normalsize,labelfont=sf,textfont=sf]{subfig}
\usepackage{textcomp}
\usepackage{stfloats}
\usepackage{url}
\usepackage{verbatim}
\usepackage{graphicx}
\usepackage{cite}

\usepackage{siunitx}
\usepackage{multirow} 
\usepackage{booktabs} 
\usepackage{xcolor}

\definecolor{darkgreen}{rgb}{0.0, 0.5, 0.0}
\definecolor{darkorange}{rgb}{0.5, 0.3, 0.1}
\definecolor{orange}{rgb}{1,0.5,0}

\newcommand{\nrx}
{N^{\text{Rx}}}

\newcommand{\ntx}
{N^{\text{Tx}}}

\title{Clustering Guided Residual Neural Networks for Multi-Tx Localization in Molecular Communications}

\author{Ali Sonmez$^*$, Erencem Ozbey$^*$, Efe Feyzi Mantaroglu, H. Birkan Yilmaz
    \thanks{A. Sonmez, E. Ozbey, E. F. Mantaroglu, and H. B. Yilmaz are with NETLAB, Department of Computer Engineering, Bogazici University, Istanbul, 34342, Turkiye (e-mail:birkan.yilmaz@bogazici.edu.tr)}% <-this % stops a space
\thanks{$^*$First and the second author contributed equally.}% <-this % stops a space
}

\begin{document}

\maketitle

\begin{abstract}
Transmitter localization in Molecular Communication via Diffusion %(MCvD) 
is a critical topic with many applications. However, accurate localization of multiple transmitters is a challenging problem due to the stochastic nature of diffusion and overlapping molecule distributions at the receiver surface. To address these issues, we introduce clustering-based centroid correction methods that enhance robustness against %noise,
density variations, and outliers. In addition, we propose two clustering-guided Residual Neural Networks, namely \textit{AngleNN} for direction refinement and \textit{SizeNN} for cluster size estimation. Experimental results show that both approaches provide significant improvements with reducing localization error between 69\% (2-Tx) and 43\% (4-Tx) compared to the K-means. 
%%%%%%%%%%%%%%%%%%%%%%%%%%%%%%%%%%%%%%%%%%%%%%%%%%%    
%

%We believe that these improvements have important implications for biomedical applications such as tumor detection, where accurate localization of multiple molecular sources is essential. 
\end{abstract}


%%%%%%%%%%%%%%%%%%%%%%%%%%%%%%%%%%%%%%%%%%%%%%%%%
%%%%%%%%%%%%%%%%%%%%%%%%%%%%%%%%%%%%%%%%%%%%%%%%%
%%%%%%%%%%%%%%%%%%%%%%%%%%%%%%%%%%%%%%%%%%%%%%%%%
%%%%%%%%%%%%%%%%%%%%%%%%%%%%%%%%%%%%%%%%%%%%%%%%%
\section{Introduction}

Molecular Communication via Diffusion (MCvD) is the process of conveying information by releasing molecules into a diffusive medium from a transmitter (Tx) node. The molecules exhibit Brownian motion in the environment~\cite{Kadloor2012_BrownMotion}, and a receiver (Rx) node attempts to decode the received molecules over time according to a chosen modulation scheme~\cite{kuran2020survey}.

MCvD occurs naturally in biological systems. For example, chemical signaling between axons and dendrites in neural communication is an example of this paradigm. Inspired by nature, MCvD has been widely investigated for applications such as in-body health monitoring, drug delivery, and tumor cell detection~\cite{nakano2012mcvd}. For reliable communication in these tasks, Rx needs to know both the number of emitted molecules and the position of the Tx, since the total number of received molecules, $\nrx$, is strongly affected by these parameters. In particular, the expected received molecule count is given by
\begin{align}
    E[\nrx] = \ntx \times F_{\text{hit}}(t,d,r) 
\end{align}
where $\ntx$ is the number of emitted molecules, and $F_{hit}(t,d,r)$ is the probability that a single molecule reaches the Rx until time $t$, given Tx–Rx center distance $d$, Rx radius $r$, and diffusion coefficient $D$.

In this work, Txs are modeled as ideal point sources, and the receiver is a perfectly absorbing spherical node. In such a scenario, $F_{hit}(t,d,r)$ is calculated as follows:
\begin{align}
    F_{\text{hit}}(t,d,r) = \frac{r}{d} \, \mathrm{erfc}\!\left(\frac{d-r}{\sqrt{4Dt}}\right)
\end{align}

Since the parameters ($r, D, t$) are known, the probability of a molecule to be received can be obtained directly when the distance $d$ is known. Similarly, $d$ can be obtained if $F_{\text{hit}}$ is known. This reduces the Tx localization problem to finding the direction vector between Rx and Tx. For a single transmitter, this direction can be estimated simply as the centroid of received molecules. 
Distance estimation~\cite{boundsdistance, distanceEstimation}, and Single-Tx localization are well investigated in the literature~\cite{nanolocal, localizeRing, localizelevenberg}, while Multi-Tx localization remains under-researched.
\begin{figure}
    \centering
    \includegraphics[width=0.92\columnwidth,keepaspectratio]{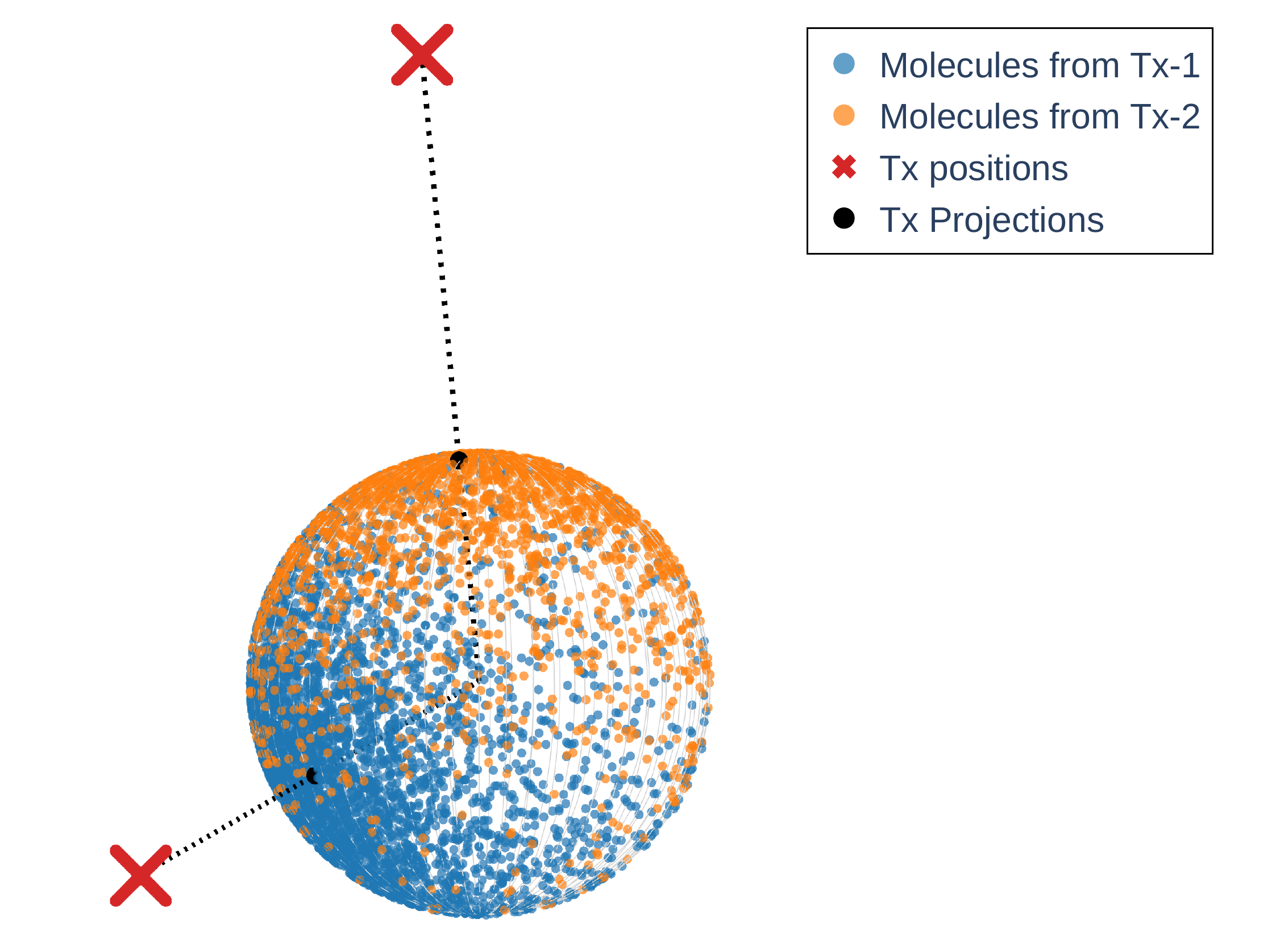}
    \caption{A Multi-Tx system where received molecules are illustrated. Emitted molecules from different point-source Tx's create overlapping distributions on the spherical Rx, making the separation and localization more challenging.}
    \label{fig:multiTx}
\end{figure}

Compared to Single-Tx scenarios, Multi-Tx localization is significantly more challenging. The overlap of molecule distributions on the Rx surface makes it difficult to assign molecules to their original transmitters (clusters), both in terms of direction estimation and size ($\nrx$) estimation. Previous work~\cite{multitx} on the topic has shown that clustering methods such as K-means, Gaussian Mixture Models (GMM) \cite{gmm}, and Dirichlet Mixture Models (DiM) \cite{dirichletMM} can be used for Multi-Tx separation, with K-means performing best. However, standard K-means suffer from two major limitations:
\begin{itemize}
    \item It ignores density variations in the molecular distribution.
    \item It is sensitive to noise and outliers caused by random Brownian motion and inter-transmitter interference.
\end{itemize}

To overcome these drawbacks, we propose K-means based centroid correction strategies and Artificial Neural Network (ANN) based refinement techniques. The main contributions of this work are as follows:
\begin{itemize}
    \item \textbf{Clustering-based improvements.} We introduce K-means–based methods that incorporate density awareness and robustness to outliers, resulting in more accurate centroid estimation for clusters on the Rx surface.
    
    \item \textbf{RNN based refinement.} We design two Residual Neural Networks (RNNs): AngleNN for direction, and SizeNN for cluster size correction to improve localization.

    \item \textbf{Performance gains.} Our  AngleNN + SizeNN refinement reduces percentage localization error of K-means by \textbf{69\%} for 2-Tx, \textbf{58\%} for 3-Tx, and \textbf{43\%} for 4-Tx scenarios. 
\end{itemize}

%%%%%%%%%%%%%%%%%%%%%%%%%%%%%%%%%%%%%%%%%%%%%%%%%
%%%%%%%%%%%%%%%%%%%%%%%%%%%%%%%%%%%%%%%%%%%%%%%%%
%%%%%%%%%%%%%%%%%%%%%%%%%%%%%%%%%%%%%%%%%%%%%%%%%
%%%%%%%%%%%%%%%%%%%%%%%%%%%%%%%%%%%%%%%%%%%%%%%%%

\section{System Model}

We consider an MCvD setup with multiple point Txs and a spherical absorbing Rx where the number of Txs is known. The Rx is assumed to have the perfect capability to sense the exact positions of the received molecules on the surface. 

In this work, the MCvD channel is simulated using a particle-based simulator that captures molecular diffusion under Brownian motion. The simulator incorporates both physical and environmental parameters of the medium, including the following: transmitter configuration (transmitters are randomly placed at distances ranging between \SI{5}{\micro\meter} and \SI{15}{\micro\meter} from the center of the receiver), receiver with radius \SI{5}{\micro\meter}, diffusion coefficient $D = \SI{79.4}{\micro\meter^2/\second}$. Each transmitter emits \SI{10000}{} molecules at the beginning of each simulation and the system is simulated with a step time of \SI{e-6}{\second}. 

This setup allows us to collect detailed molecule arrival statistics at the Rx, which form the basis for clustering-based localization and RNN refinement techniques.

%%%%%%%%%%%%%%%%%%%%%%%%%%%%%%%%%%%%%%%%%%%%%%%%%
%%%%%%%%%%%%%%%%%%%%%%%%%%%%%%%%%%%%%%%%%%%%%%%%%
%%%%%%%%%%%%%%%%%%%%%%%%%%%%%%%%%%%%%%%%%%%%%%%%%
%%%%%%%%%%%%%%%%%%%%%%%%%%%%%%%%%%%%%%%%%%%%%%%%%
\section{Methodology}

In this letter, we propose two methodologies for Multi-Tx localization: (i) algorithmic correction of cluster centers obtained from K-means clustering, and (ii) RNN-based refinement that leverage K-means cluster statistics as input features.

\subsection{Algorithmic Correction of Cluster Centers}
In the first stage, the particle arrival points are grouped using the K-means algorithm, providing a preliminary estimate of transmitter-related clusters. To improve centroid estimation, we investigate three post-processing methods:

\subsubsection*{\textbf{Minimum Covariance Determinant (MinCovDet)}} This method identifies a robust subset of inliers within each K-means cluster using the Minimum Covariance Determinant estimator \cite{mincovdet}. The cluster center is then redefined as the robust mean of these inliers, reducing the influence of outliers.
    
\subsubsection*{\textbf{Density-Weighted Centroid}} After obtaining the final clusters, instead of assigning equal importance to all cluster members, each point is weighted according to its estimated local density, calculated from the mean distance to its nearest $k$ neighbors. Points in dense regions receive higher weights, ensuring that the centroid reflects the true cluster core. The revised centroid is then given by
\begin{align}
    \mathbf{c}_{\text{dw}} = 
    \frac{\sum_{j=1}^{n} w_j \mathbf{x}_j}{\sum_{j=1}^{n} w_j}
\end{align}
where $\mathbf{x}_j$ is the $j$-th point in the cluster, $w_j$ is its density-based weight, and $n$ is the number of points in the cluster. This adjustment shifts the centroid toward the densest region of the cluster, which mitigates the influence of noise and outliers.
    
\subsubsection*{\textbf{Density-Weighted MinCovDet (Density MinCovDet)}} This hybrid approach first applies MinCovDet to isolate robust inliers and then computes a density-weighted centroid over the inlier set. This combines the outlier resistance of MinCovDet with the density sensitivity of weighted centroids.

These three correction techniques serve as algorithmic alternatives to the baseline K-means, producing more accurate and robust cluster estimates.

\subsection{Refinement with Residual Networks}
Although algorithmic corrections improve centroid accuracy, they are limited in terms of capturing the full complexity of molecular diffusion. To overcome these limitations, we introduce two clustering-guided RNNs designed to refine the estimates obtained from K-means clustering. Specifically, the goal is to improve both the \emph{directions} of transmitter locations and the \emph{sizes} of the associated molecular clusters.

We consider a multi-transmitter scenario with $K$ transmitters, where K-means algorithm provides $K$ cluster centers and their corresponding sizes. Each cluster is represented by four features: size and the three Cartesian coordinates of its centroid. These $4K$ values form the input to our RNN models.

Both proposed models, \textbf{AngleNN} and \textbf{SizeNN}, share similar architectures with different output dimensions:  
\begin{itemize}
    \item An input layer of dimension $4K$   
    \item A hidden stem projecting input to 256 dimensions, followed by six stacked residual blocks, each consisting of fully connected layers, ReLU activations, and dropout. %for regularization.  
    \item A head that reduces the 256-dimensional hidden representation to the desired output dimension.  
\end{itemize}

AngleNN predicts Tx \emph{directions}. Its output dimension is $3K$, corresponding to the corrected 3D vectors pointing from the Rx center to each of the $K$ Tx nodes. The residual design ensures that the network learns corrections relative to the K-means estimates rather than predicting directions from scratch.  

SizeNN predicts cluster \emph{sizes}. Its output dimension is $K$, corresponding to the effective number of molecules received from each transmitter. By leveraging the same residual architecture, SizeNN systematically adjusts the K-means size estimates to reduce the error.  

By combining these two refinements, AngleNN and SizeNN provide complementary corrections: AngleNN improves the accuracy of directional localization, while SizeNN enhances the estimation of molecule counts. Together, they deliver substantial improvements in Multi-Tx localization.

%%%%%%%%%%%%%%%%%%%%%%%%%%%%%%%%%%%%%%%%%%%%%%%%%
%%%%%%%%%%%%%%%%%%%%%%%%%%%%%%%%%%%%%%%%%%%%%%%%%
%%%%%%%%%%%%%%%%%%%%%%%%%%%%%%%%%%%%%%%%%%%%%%%%%
%%%%%%%%%%%%%%%%%%%%%%%%%%%%%%%%%%%%%%%%%%%%%%%%%
\section{Performance Evaluation}

\begin{figure*}
    \centering
    \includegraphics[width=1\linewidth]{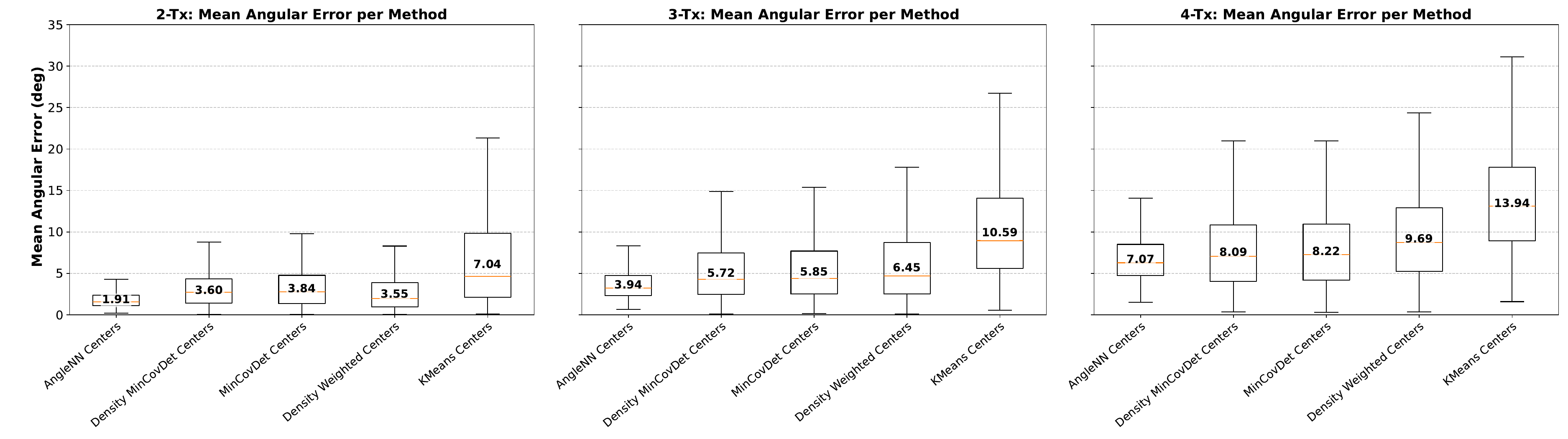}
    \caption{Mean angular error (in degrees) for different methods under 2-Tx, 3-Tx and 4-Tx scenarios.}
    \label{fig:boxplot}
\end{figure*}
\begin{figure*}
    \centering
    \includegraphics[width=1\linewidth]{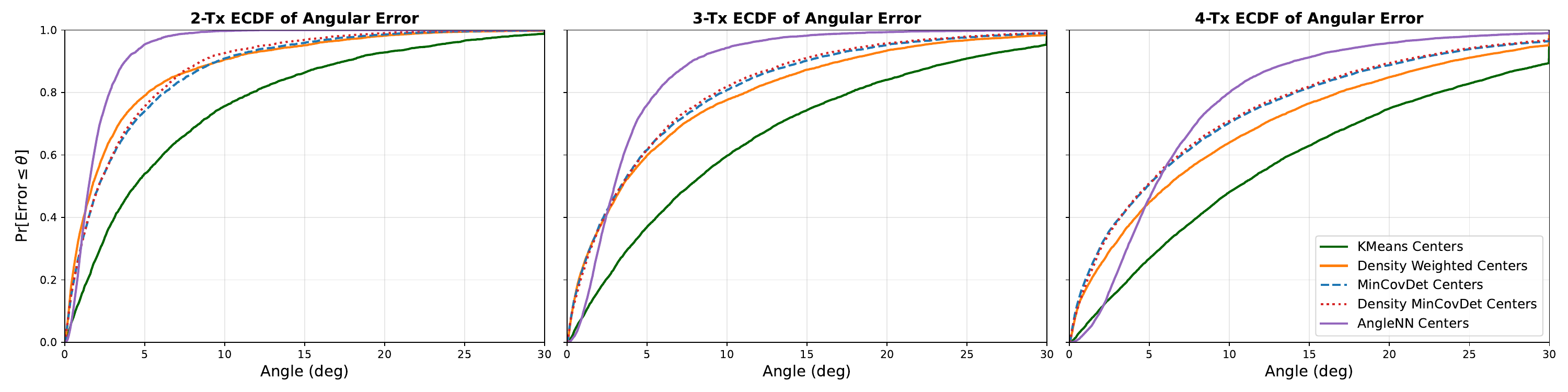}
    \caption{Empirical cumulative distribution functions (ECDFs) of angular errors in 2-Tx, 3-Tx, and 4-Tx scenarios.}
    \label{fig:cumulative}
\end{figure*}

Figure \ref{fig:boxplot} compares the mean angular error for 2-Tx, 3-Tx, and 4-Tx scenarios. Lower values indicate a more accurate estimation of transmitter directions. As expected, K-means exhibits the highest error due to its sensitivity to noise and outliers. In addition, error distributions are wide and unstable, frequently extending beyond \SI{20}{\degree} in the 2-Tx case, \SI{25}{\degree} in the 3-Tx case, and \SI{30}{\degree} for 4-Tx.

The Density-Weighted and MinCovDet methods reduce both mean error and variance compared to K-means. As a hybrid technique, the Density MinCovDet, achieves the best clustering-based performance. However, noticeable error spread remains, particularly under 3-Tx and 4-Tx conditions, where molecule overlap increases on the Rx surface.

In contrast, the proposed AngleNN consistently achieves the lowest angular errors and the most stable error distributions. For 2-Tx, AngleNN maintains estimates tightly concentrated below \SI{5}{\degree}, with very few outliers. For 3-Tx, although all methods show increased difficulty and broader distributions, AngleNN still limits errors to below \SI{10}{\degree}.

These results demonstrate that while robust clustering techniques improve over K-means, RNN-based refinement is essential for reducing both average angular error and variability.

Figure \ref{fig:cumulative} presents the empirical cumulative distribution functions of the proposed methods to provide a general perspective on localization accuracy, clearly illustrating the advantage of AngleNN. In all 2-Tx, 3-Tx, and 4-Tx cases, the AngleNN curve rises steeply, whereas K-means technique lags far behind with a much slower climb. The separation between the curves is especially pronounced in the lower-angle region, highlighting that AngleNN consistently delivers highly precise estimates. Although Density MinCovDet Centers and Density Weighted Centers improve upon K-means by shifting their curves upward, they remain below AngleNN and exhibit weaker performance. The ECDF plots thus reinforce that AngleNN not only reduces average errors but also ensures that the vast majority of localizations fall within tight angular bounds, demonstrating strong reliability across scenarios.

\begin{figure*}
    \centering
    \includegraphics[width=1\linewidth]{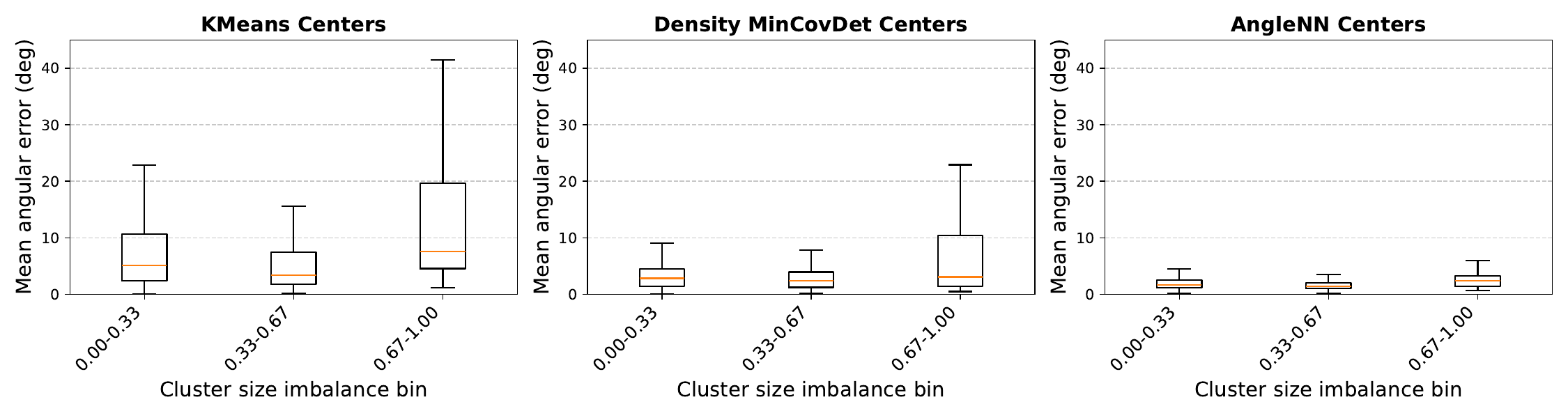}
    \caption{Localization performance under varying cluster size imbalance for 2-Tx.}
    \label{fig:size_imbalance}
\end{figure*}
In Figure \ref{fig:size_imbalance}, we analyze localization performance with respect to the cluster size imbalance obtained from K-means outputs. Although each transmitter emits the same number of molecules, their different distances from the receiver result in true clusters that are inherently imbalanced —closer transmitters contribute more molecules. K-means reflects this imbalance in its estimated clusters, which we quantify using the imbalance metric defined below. Formally, for two clusters of sizes $n_1$ and $n_2$, the imbalance is defined as 
\begin{align}
    \text{Imbalance}(n_1,n_2) = \frac{|n_1 - n_2|}{n_1 + n_2},
\end{align}  
which ranges from $0$ (perfectly balanced) to $1$ (maximally imbalanced). For more than two clusters, the reported imbalance corresponds to the mean pairwise imbalance.  

As imbalance increases, so does error variability. K-means is the most sensitive, with median errors rising sharply in the high-imbalance regime, sometimes exceeding $40^\circ$. AngleNN demonstrates remarkable robustness, maintaining low errors and only slight variability across all cluster imbalance levels. RNN-based refinement methods outperform clustering approaches when molecule counts are highly uneven.

To compare the percentage error across different methods, we use  Mean Absolute Percentage Error (MAPE). Formally, MAPE for two vectors is defined as
\begin{align}
    \text{MAPE} = 100 \times \frac{1}{N} \sum_{i=1}^{N} 
    \frac{\|\mathbf{y}_i - \hat{\mathbf{y}}_i\|}{\|\mathbf{y}_i\|},
\end{align}
where $\hat{\mathbf{y}}_i$ and $\mathbf{y}_i$ denote the predicted and actual position vectors of the $i$-th Tx and $N$ represents the total number of transmitter instances in all experiments.

\begin{table}[t]
  \centering
  \caption{Mean Angular Error Per Method for Center Estimation}
  \label{tab:mean_angular_diff_per_method}
  \begin{tabular}{lccc}
    \toprule
    \textbf{Method} & \textbf{2 Tx} & \textbf{3 Tx} & \textbf{4 Tx} \\
    \midrule
    AngleNN                 & \textbf{\SI{1.91}{\degree}} & \textbf{\SI{3.94}{\degree}} & \textbf{\SI{7.07}{\degree}} \\
    Density MinCovDet   & \SI{3.60}{\degree} & \SI{5.72}{\degree} & \SI{8.09}{\degree} \\
    MinCovDet           & \SI{3.84}{\degree} & \SI{5.85}{\degree} & \SI{8.22}{\degree} \\
    Density             & \SI{3.55}{\degree} & \SI{6.45}{\degree} & \SI{9.69}{\degree} \\
    K-means              & \SI{7.04}{\degree} & \SI{10.59}{\degree} & \SI{13.94}{\degree} \\
    GMM              & \SI{14.24}{\degree} & \SI{17.43}{\degree} & \SI{19.21}{\degree} \\
    \bottomrule
  \end{tabular}
\end{table}
Table \ref{tab:mean_angular_diff_per_method} summarizes the mean angular errors for each method. AngleNN achieves the lowest errors across all scenarios and substantially improves over K-means (the best method in literature), with reductions of 73\%, 63\%, and 49\% in the 2-Tx, 3-Tx, and 4-Tx cases, respectively. Among clustering-only methods, Density-MinCovDet and MinCovDet are better than density weighting. These results demonstrate that for high accuracy RNN-based refinement is essential.

\begin{table}[ht]
  \centering
  \caption{Performance of Cluster Size Estimation }
  \label{tab:size_estimation_performance}
  \begin{tabular}{lcccccc}
    \toprule
    \multirow{2}{*}{\textbf{Method}} & \multicolumn{3}{c}{\textbf{MAPE (\%)}} & \multicolumn{3}{c}{\textbf{RMSE}} \\
    \cmidrule(lr){2-4} \cmidrule(lr){5-7}
     & \textbf{2 Tx} & \textbf{3 Tx} & \textbf{4 Tx} & \textbf{2 Tx} & \textbf{3 Tx} & \textbf{4 Tx} \\
    \midrule
    SizeNN   & \textbf{3.66} & \textbf{8.76} & \textbf{14.88} & \textbf{357.04} & \textbf{752.99} & \textbf{954.12} \\
    K-means  & 7.58 & 12.69 & 17.84 & 670.77 & 1032.33 & 1249.19 \\
    \bottomrule
  \end{tabular}
\end{table}
Table \ref{tab:size_estimation_performance} summarizes the accuracy of cluster size estimation in terms of MAPE and RMSE. Across all transmitter scenarios, SizeNN consistently outperforms K-means, achieving error reductions of up to 52\% in MAPE and 47\% in RMSE.

\begin{table}[t]
  \centering
  \caption{MAPE (\%) Per Method for Multi-Tx Localization }
  \label{tab:mape_per_method}
  \begin{tabular}{llccc}
    \toprule
    \textbf{Center Estimation} & \textbf{Size Estimation} & \textbf{2 Tx} & \textbf{3 Tx} & \textbf{4 Tx} \\
    \midrule
    AngleNN & SizeNN    & \textbf{3.99} & \textbf{8.12} & \textbf{14.30} \\
    AngleNN & K-means           & 5.06 & 8.89 & 14.53 \\
 
    Density MinCovDet & K-means  & 7.52 & 11.54 & 16.05 \\
    MinCovDet &  K-means        & 7.90 & 11.76 & 16.27 \\
    Density  &   K-means        & 7.36 & 12.61 & 18.44 \\
    \midrule
    K-means   & SizeNN          & 12.42 & 18.75 & 25.00 \\
    K-means   &   K-means        & 12.86 & 19.12 & 25.14 \\
    \bottomrule
  \end{tabular}
\end{table}
Table \ref{tab:mape_per_method} summarizes the MAPE values for localization across different methods. The joint AngleNN + SizeNN refinement consistently outperforms all baselines, reducing errors by 69\%, 58\%, and 43\% compared to K-means in 2-Tx, 3-Tx, and 4-Tx scenarios, respectively. AngleNN alone also performs strongly, especially in higher Tx cases where its accuracy approaches that of the combined model, but the addition of SizeNN provides clear benefits in all settings.  SizeNN on its own improves size estimation relative to K-means but remains less effective than AngleNN, highlighting that direction accuracy is the primary driver of overall performance. These results emphasize the complementary strengths of AngleNN and SizeNN, where direction refinement and size correction together deliver the most reliable localization outcomes.

%%%%%%%%%%%%%%%%%%%%%%%%%%%%%%%%%%%%%%%%%%%%%%%%%
%%%%%%%%%%%%%%%%%%%%%%%%%%%%%%%%%%%%%%%%%%%%%%%%%
%%%%%%%%%%%%%%%%%%%%%%%%%%%%%%%%%%%%%%%%%%%%%%%%%
%%%%%%%%%%%%%%%%%%%%%%%%%%%%%%%%%%%%%%%%%%%%%%%%%
\section{Conclusion}

This paper addressed the challenging problem of multiple transmitter localization in an MCvD system. We first analyzed the limitations of K-means clustering, which suffers from sensitivity to noise, outliers, and cluster size imbalance. To mitigate these issues, we introduced three correction strategies based on density weighting and robust covariance estimation, and further proposed two clustering-guided RNNs, AngleNN and SizeNN, for refinement of direction and size estimation.  

Our results demonstrated that AngleNN  achieved the lowest angular errors, reducing localization error by up to \textbf{73}\% compared to K-means, while SizeNN substantially improved cluster size estimation accuracy. When combined, AngleNN and SizeNN provided the most reliable localization, lowering the mean absolute percentage error by \textbf{69}\%, \textbf{58}\%, and \textbf{43}\% for 2-Tx, 3-Tx, and 4-Tx scenarios, respectively.

\begin{comment}

\end{comment}

\bibliographystyle{IEEEtran} 
\bibliography{refs}

\end{document}